\documentclass[runningheads]{llncs}
\usepackage{graphicx}
\usepackage{amssymb}
\usepackage{booktabs}
\usepackage{array}
\usepackage[font=small]{caption}
\usepackage{subcaption}
\usepackage{float}
\usepackage{multirow}

\usepackage{url,hyperref}
\usepackage[ruled,vlined]{algorithm2e}

\begin{document}
\title{GrabQC: Graph based Query Contextualization for automated ICD coding}

\titlerunning{GrabQC: Graph based Query Contextualization for automated ICD coding}

\author{Jeshuren Chelladurai\inst{1,2} \and
Sudarsun Santhiappan\inst{1} \and
Balaraman Ravindran\inst{1,2}}

\authorrunning{J. Chelladurai et al.}
%
\institute{Dept. of Computer Science and Engineering, Indian Institute of Technology Madras \and
Robert Bosch Centre for Data Science and AI \\
\email{\{jeshuren,sudarsun,ravi\}@cse.iitm.ac.in}}

\maketitle              
\begin{abstract}

Automated medical coding is a process of codifying clinical notes to appropriate diagnosis and procedure codes automatically from the standard taxonomies such as ICD (International Classification of Diseases) and CPT (Current Procedure Terminology). The manual coding process involves the identification of entities from the clinical notes followed by querying a commercial or non-commercial medical codes Information Retrieval (IR) system that follows the Centre for Medicare and Medicaid Services (CMS) guidelines. We propose to automate this manual process by automatically constructing a query for the IR system using the entities auto-extracted from the clinical notes. We propose \textbf{GrabQC}, a \textbf{Gra}ph \textbf{b}ased \textbf{Q}uery \textbf{C}ontextualization method that automatically extracts queries from the clinical text, contextualizes the queries using a Graph Neural Network (GNN) model and obtains the ICD Codes using an external IR system. We also propose a method for labelling the dataset for training the model. We perform experiments on two datasets of clinical text in three different setups to assert the effectiveness of our approach. The experimental results show that our proposed method is better than the compared baselines in all three settings.

\end{abstract}
\section{Introduction}

Automated medical coding is a research direction of great interest to the healthcare industry \cite{article:25-billion,inproceedings:25-billion-2}, especially in the Revenue Cycle Management (RCM) space, as a solution to the traditional human-powered coding limitations.  Medical coding is a process of codifying diagnoses, conditions, symptoms, procedures, techniques, the equipment described in a medical chart or a clinical note of a patient.  The codifying process involves mapping the medical concepts in context to one or more accurate codes from the standard taxonomies such as ICD (International Classification of Diseases) and CPT (Current Procedure Terminology). The ICD taxonomy is a hierarchy of diagnostic codes maintained by the World Health Organisation. Medical coders are trained professionals who study a medical chart and assign appropriate codes based on their interpretation. The most significant drawbacks in manual coding are its Turn-Around Time (TAT), typically 24-48 hours, and the inability to scale to large volumes of data.  Automatic medical coding addresses both problems by applying AI and Natural Language Understanding (NLU) by mimicking and automating the manual coding process. 

\begin{figure}[t]
    \centering
    \includegraphics[scale=0.3]{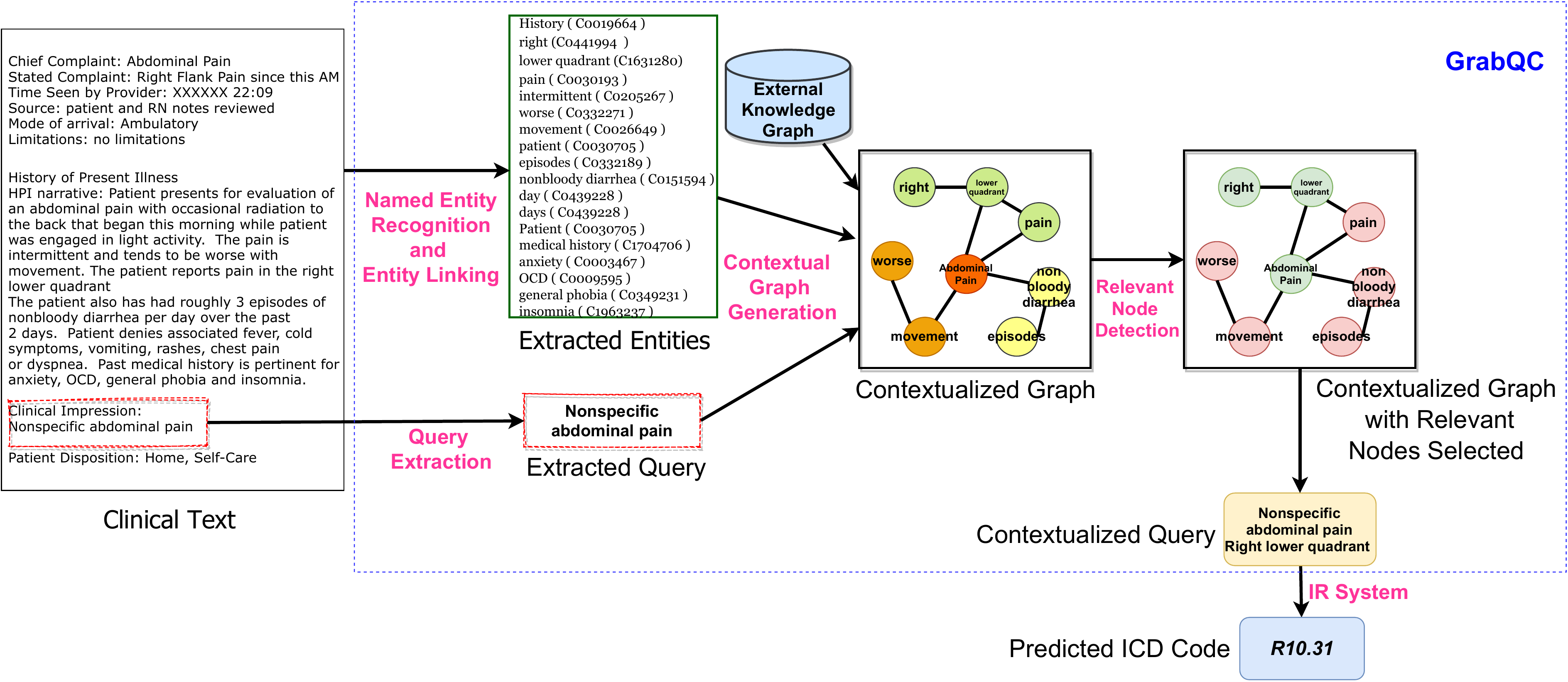}
    \caption{The figure describes the overall pipeline of our proposed method.}
    \label{fig:our-method}
\end{figure}


The problem of automating the assignment of ICD codes to clinical notes is challenging due to several factors, such as the lack of consistent document structure, variability in physicians' writing style, choice of vocabularies to represent a medical concept, non-explicit narratives, typographic and OCR conversion errors.  Several approaches to solve these problems are available in the literature such as methods based on Deep Learning \cite{inproceedings:25-billion-2,DUARTE201864:DLforICD,article:DLforICD2,mullenbach-etal-2018-explainable:caml,unknown:bertXML}, Knowledge Bases \cite{10.3389/fbioe.2020.00867:KGforICD} and Extractive Text Summarisation \cite{inproceedings:text-summarization-for-ICD}. The most recent works treat this problem as a multi-label classification problem and solve them using various deep learning model architectures. Although approaches based on deep learning  greatly reduce the manual labor required in the feature engineering process, there are certain challenges in applying these to the medical coding task: 
\begin{itemize}
    \item Lack of explainability
    \item Requirement of large amounts of labelled data
    \item Large label space (ICD9 $\sim$14,000 codes, ICD10 $\sim$72,000 codes)
\end{itemize}

There have been several attempts to address these challenges, for e.g., using an attention mechanism \cite{mullenbach-etal-2018-explainable:caml,li2020multirescnn,inproceedings:ml-5} ,  transfer learning \cite{unknown:bertXML} and extreme classification \cite{Bhatia16,unknown:bertXML}. However, much more needs to be done to develop truly satisfactory deployable systems. In this work, we focus on the question of operating with large label space. We study how a medical coder arrives at the final ICD codes for a given clinical note or medical chart.  We observe that the medical coders study entities from different subsections of a document, such as Chief Complaint, Procedure, Impression, Diagnosis, Findings, etc., to construct evidence for every ICD code.  We also observe that the medical coders use several commercial and non-commercial Information Retrieval (IR) tools such as Optum EncoderPro 
, AAPC Codify
and open tools such as CDC's ICD-CM-10 Browser Tool
for associating the entities to relevant ICD codes. We propose a solution for automated ICD coding by automatically constructing a contextually enhanced text query containing entities from the clinical notes along with an existing information retrieval system.  Fig. \ref{fig:our-method} depicts our method, which extracts an accurate entities based query, with which the relevant ICD codes could be fetched by querying an ICD IR system. Our method provides explainability to the retrieved codes in terms of the contextual entities, usually lacking in the end-to-end DL based solutions.


We propose GrabQC, a Graph-based Query contextualization method, alongside an existing Information Retrieval system to automatically assign ICD codes to clinical notes and medical records. The overall architecture of our proposed method, which consists of four essential modules, is shown in Fig. \ref{fig:our-method}. The first module (Section \ref{Named Entity Recognition}) extracts all the data elements (entities) from the clinical notes along with their respective types such as condition, body part, symptom, drug, technique, procedure, etc.  The second module (Section \ref{Query Extraction}) extracts the primary diagnosis available typically under the Chief Complaint section of a medical chart.  The third module (Section \ref{Generation of Contextualized Graph}) constructs a graph from the entities enriched by a pre-constructed external Knowledge Base. The fourth module (Section \ref{model}) prunes the constructed graph based on relevance to the clinical note concepts.  We then construct the contextualized query for the integrated IR system to fetch the relevant ICD codes.


The main contributions in this work are as follows:

\begin{itemize}
    \item \textit{GrabQC}, a Graph-based Query Contextualization Module to extract and generate contextually enriched queries from clinical notes to query an IR system.
    \item A Graph Neural Network (GNN) model to filter relevant nodes in a graph to contextualize the concepts.
    \item A distant supervised method to generate labelled dataset for training the GNN model for relevant node detection.
\end{itemize}

The rest of this paper is organised as follows. In Section \ref{Related Work}, we provide a summary of the other approaches in the literature that solve the problem of ICD coding. We describe the proposed GrabQC module in Section \ref{GrabQC}. We present our experimental setup and the results in Section \ref{Experiments} and \ref{Results} respectively. We finally give concluding remarks and possible future directions of our work in Section \ref{Conclusion}. 

\section{Related Work} \label{Related Work}

Automatic ICD coding is a challenging problem, and works in literature have focused on solving this problem using two broad paradigms \cite{inproceedings:IR}, Information Retrieval (IR) and Machine Learning (ML). Our current work is a hybrid approach that uses both paradigms. We first use ML to contextualize the query based on the extracted entities, and then use the contextualized query for inquiring the IR system. Our work is a special case of query expansion, where we use the source document to expand by contextualizing the extracted query.

\paragraph{Information Retrieval:} Rizzo et al. \cite{inproceedings:IR} presented an incremental approach to the ICD coding problem through top-K retrieval. They also used the Transfer Learning
techniques to expand the skewed dataset. The authors also established experimentally that the performance of ML methods decreases as the number of classes increases. Park et al. \cite{article-IR2} used IR to map free-text disease descriptions to ICD-10 codes.

\paragraph{Machine/Deep Learning:} Automatic ICD coding problem has also been solved by framing it as a multi-label classification problem \cite{inproceedings:25-billion-2,DUARTE201864:DLforICD,article:DLforICD2,article:ml-2,inproceedings:ml-3,mullenbach-etal-2018-explainable:caml,li2020multirescnn,inproceedings:ml-5}. The past works can be broadly classified as belonging to one of the three categories, ML feature engineering \cite{10.1145/243199.243276:ml-feature}, Deep representation learning architectures \cite{mullenbach-etal-2018-explainable:caml,article:DLforICD2,DUARTE201864:DLforICD,inproceedings:ml-5} and addition of additional information, such as the hierarchy of ICD codes \cite{inproceedings:additional-info}, label descriptions \cite{unknown:bertXML,mullenbach-etal-2018-explainable:caml}. There are also research works that solve these problem on other forms of clinical texts, such as the Death certificates \cite{article:ml-2} and Radiology reports \cite{inproceedings:ml-3}. 

\paragraph{Graph based DL for Automated ICD Coding:} Very few works have used Graphs in solving the problem of ICD coding. Graph convolutional neural networks \cite{kipf2017semi:gcn} was used for regularization of the model to capture the hierarchical relationships among the codes. Fei et al. \cite{10.3389/fbioe.2020.00867:KGforICD} proposed a method which uses graph representation learning to incorporate information from the Freebase Knowledge Base about the diseases in the clinical text. The learned representation was used along with a Multi-CNN representation of the clinical texts to predict the ICD codes.

\section{GrabQC - Our Proposed Approach} \label{GrabQC}

Automatic ICD coding is the task of mapping a clinical note $\mathcal{D}$ into a subset of the ICD codes $\mathcal{Y}$. We propose GrabQC, a graph-based Query Contextualization method to solve this problem using a hybrid of Deep Learning and Information Retrieval approaches. We will now discuss this proposed method in detail.

\subsection{Named Entity Recognition and Entity Linking} \label{Named Entity Recognition}

A semi-structured clinical note $\mathcal{D}$ consists of a set of subsections $\mathcal{T}$, where each section comprises one or more sentences. The initial step is to identify interesting entities, which would serve as building blocks of the contextualized graph. We use a Named Entity Recognition (NER) module to extract a set of entities, $ \mathcal{E} = \{ e_1, e_2, \dots , e_m \}$ where $e_{i}$ can be a single word or a span of words from $\mathcal{D}$. To obtain more information about the extracted entities, we need to associate them with a knowledge base $\mathcal{B}$. An entity linker takes as input the sentence and the extracted entity and outputs a linked entity from the Knowledge base $\mathcal{B}$. The extracted entity set $\mathcal{E}$ is linked with $\mathcal{B}$ using the entity linker to produce the set of linked entities $ \mathcal{L} = \{ l_1, l_2, \dots , l_m \}$.

\subsection{Query Extraction} \label{Query Extraction}

The physician lists the diagnoses of the patient in a pre-defined subsection of the clinical note. This subsection can be extracted using rules and regex-based methods. Each item in the extracted list of diagnoses constitute the initial query set $ \mathcal{Q} = \{ q_1, q_2, \dots , q_n \}$, where $n$ is the number of queries/diagnoses. We also extract entities from the queries in $\mathcal{Q}$ and link them with the $\mathcal{B}$ to produce a set of linked entities $\mathcal{P}_i$, for each $q_i$.

\subsection{Generation of Contextual Graph} \label{Generation of Contextualized Graph}

From the methods described in Section \ref{Named Entity Recognition} and \ref{Query Extraction}, we convert the clinical note $\mathcal{D}$ into a set of queries $\mathcal{Q}$, set of linked entities $\mathcal{L}$ and set of query-linked entities $\mathcal{P}_i, \forall q_{i} \in \mathcal{Q}$. We generate a Contextual Graph $\mathcal{G}_{i}$, for each query $q_{i} \in \mathcal{Q}$. We construct a set of matched entities $V_{i}$, with the entities in $\mathcal{P}_{i}$ that have a simple path with $\mathcal{L}$ in a Knowledge Base  ($\mathcal{B}$). We construct the edge set of $\mathcal{G}_{i}$, $E_{i}$ by adding an edge $(p_{i}, l_{j})$, if the path $(p_{i}, \cdots, l_{j}) \in \mathcal{B}$. For every matched entity, we add an edge between the matched entity and other linked entities in the same sentence. Repeating this process, we obtain the contextual graph $\mathcal{G}_{i} = (V_{i},E_{i}) , \forall q_{i} \in \mathcal{Q}$.

\subsection{Relevant Node Detection} \label{model}

We frame the problem of detecting relevant nodes in a Contextual Graph ($\mathcal{G}$) as a supervised node classification problem. Let $\mathcal{G}$ be the contextual graph of the query $q$, obtained from Section \ref{Generation of Contextualized Graph}. Here $\mathcal{G} = (V,E)$, where $V = \{n_{1}, n_{2}, \cdots, n_{N} \}$ is the set of nodes and $E \subseteq \{ V \times V $ \} is the set of edges. The structure of the $\mathcal{G}$ is denoted using an adjacency matrix,  $ A \in \mathbb{R}^{N \times N}$, where $A_{ij} = 1$ if the nodes $i$ and $j$ are connected in the $\mathcal{G}$, 0 otherwise. Given $\mathcal{G}$ and query $q$, we train a neural network which passes $\mathcal{G}$ through a Graph Neural Network(GNN) model to compute the hidden representation for all the nodes in the $\mathcal{G}$. We also obtain the representation for $q$ by averaging over the words' embeddings in the query and passing it through a dense layer. We then concatenate the individual nodes' GNN representations with the query to obtain the final representation for each node-query pair. Finally, these node-query pair representations are passed to a softmax layer to classify them as either relevant or non-relevant. We then obtain the contextual query $q_{c}$ by concatenating the query with the predicted relevant nodes' words. We will now describe each of these elements in more detail.

At the base layer of the model we have $d$-dimensional pre-trained word embeddings for each word ($v_{i}$) in the node of the contextual graph. The $d$-dimensional embeddings ($x_i$) of the ${N}$ nodes in the $\mathcal{G}$ are vertically stacked to form the initial feature matrix, $\mathbf{X} ^{(0)} = [x_1; x_2; \cdots; x_{N}] \in \mathbb{R}^{N \times d}$.  The node representations at each layer are obtained using the Graph Neural Network as described by Morris et al., \cite{Morris_Ritzert_Fey_Hamilton_Lenssen_Rattan_Grohe_2019:GNN} with $\mathbf{W}^{(\ell)}_1$ and $\mathbf{W}^{(\ell)}_2$ as the parameters in each layer. We use this variant of the GNN rather than the widely-used Graph Convolutional Networks \cite{kipf2017semi:gcn} because it preserves the central node information and omits neighbourhood normalization. To obtain the hidden representation at each layer, we perform,

\begin{equation}
    \mathbf{X}^{(\ell+1)} = \mathbf{W}^{(\ell + 1)}_1 \mathbf{X}^{(\ell)} + \mathbf{W}^{(\ell + 1)}_2 A \mathbf{X}^{(\ell)}
\end{equation}

To obtain the hidden representation for the query $q$, we average the $d$-dimensional pre-trained word embeddings($w_i$) of the words present in $q$. Then we pass the averaged word embedding through a dense layer, with $\mathbf{W}_q,\mathbf{b}_q $ as its learnable parameters. The hidden representation of the query ($\mathbf{x}^{q}$) is given by,

\begin{equation}
    \mathbf{x}^{q} = \mathbf{W}_q \left( \frac{1}{|w|} \sum_{w_i \in w} w_i \right)  + \mathbf{b}_q
\end{equation}

The input to the final classification layer is the combined representation of the node in the final layer ($k$) of the GNN and the query's hidden representation. The final prediction of the model whether a given node $n_i \in \mathcal{G}$ is relevant for the query $q$ is given by,

\begin{equation}
    Pr( n_{i} = \textit{relevant} \mid q) = \textbf{sigmoid}(\mathbf{W}_{c} [\mathbf{x}^{k}_{i}; \mathbf{x}_q] + \mathbf{b}_{c} )
\end{equation}

\begin{equation}
    \hat{y}_{i} = \mathbb{I} \left[ Pr( n_{i} = \textit{relevant} \mid q) \geq \tau \right]
\end{equation}

Here $\mathbb{I}$ is an indicator function that outputs 1 if the score is greater than the threshold ($\tau$), 0 otherwise. We train the model by optimizing the Weighted Binary Cross-Entropy Loss with $L_2$-regularization using the Adam Optimizer \cite{article:adam}.

The contextual query $q_{c}$ is given by,
\begin{equation}
    q_{c} = \{ q \cup v_{i} \mid \hat{y}_{i} =1 ,  1 \leq i \leq N \} 
\label{eqn:contextual-query}
\end{equation}

\subsection{Distant Supervised Dataset Creation}

The training of the model described in Section \ref{model} requires labels for each node in the contextual graph $\mathcal{G}$. 
Gathering relevance labels for each node in the $\mathcal{G}$ is challenging and time-consuming, as the available datasets have ICD codes only for each clinical chart. Hence, we use distant supervision \cite{10.5555/1690219.1690287:distant-supervision} for annotating the nodes in the $\mathcal{G}$. We chose to use the descriptions of the ICD codes\footnote{https://www.cms.gov/Medicare/Coding/ICD10/2018-ICD-10-CM-and-GEMs} for providing supervision to the labelling process. 

Let $\mathcal{D}$ be the clinical note and $L = \{l_1, \cdots, l_m\}$ be the manually annotated ICD-codes for $\mathcal{D}$. Let the descriptions for the ICD codes in $Y$ be $B = \{b_1, \cdots, b_m\}$. We use the methods described in Section \ref{Generation of Contextualized Graph}, to convert $\mathcal{D}$ into $\mathcal{Q} = \{q_1, \cdots,q_n\}$ and $\mathcal{G} = \{\mathcal{G}_1, \cdots,\mathcal{G}_n\}$, where $\mathcal{G}_i = (V_{i},E_{i})$ . We need to assign a relevancy label $y_{j}, \forall v_{j} \in V_{i}$ . Let $W_{j}$ be the set of words in node $v_{j} \in V_{i}$ and $l_{k}$ be the label corresponding to $q_{i}$. We label $v_{j} \in V_{i}$ to be relevant ($y_j = 1$), if the Jaccard similarity of $W_{j}$ and $b_{k}$ is greater than a threshold($t$). Otherwise, we label $v_{i}$ as non-relevant ($y_j = 0$). Repeating this process for all the nodes, we obtain the node-labelled $\mathcal{G}_i$.

\section{Experimental Setup} \label{Experiments}

In this section, we perform experiments to answer the following questions,
\begin{itemize}
    \item How does the proposed relevant node detection model perform in identifying the relevant nodes in the contextualized graph? (Section \ref{result:1})
    \item Is the query's performance generated by GrabQC better than the initial query on the same IR system? (Section \ref{result:2})
    \item How does the proposed method compare against other machine-learning-based methods in the literature? (Section \ref{result:3})
\end{itemize}

\subsection{Dataset Description}
We use anonymized versions of two clinical text datasets, a subset of the publicly available MIMIC-III dataset and a proprietary dataset (EM-clinical notes). The MIMIC-III dataset is a collection of discharge summaries of patients from the Intensive Care Unit (ICU). At the same time, the EM-clinical notes (referred to as \textit{EM}) are Electronic Health Records (EHRs) of patients visiting the Emergency Room in a hospital. Both the datasets are manually coded with the corresponding ICD diagnosis codes. The choice of dataset from different specialties help in understanding the performance of our method across them. The subset of MIMIC-III (referred to as \textit{M3s}) dataset consists of discharge summaries. The number of ICD codes and the number of extracted queries had a difference of less than 2. M3s dataset contains 7166 discharge summaries with 1934 unique ICD-9 codes, while the EM dataset has 7991 EHRs with 2027 ICD-10 codes. Since ICD-10 is the latest version used commercially and lacks a public source ICD-10 coded clinical notes dataset, we use a proprietary dataset to test our method.

\setlength{\tabcolsep}{10pt} 
\renewcommand{\arraystretch}{1.5}

\subsection{Setup and Baselines}

We follow three experimental setups to assess the performance of the proposed method. For all the experiments, we have split the datasets into train/dev/test splits. We have trained our model on the training set by tuning the hyper-parameters in the dev set. We report the results on the test set. 

\begin{figure}
  \centering
  \begin{subfigure}[b]{0.5\linewidth}
    \centering\includegraphics[scale = 0.2]{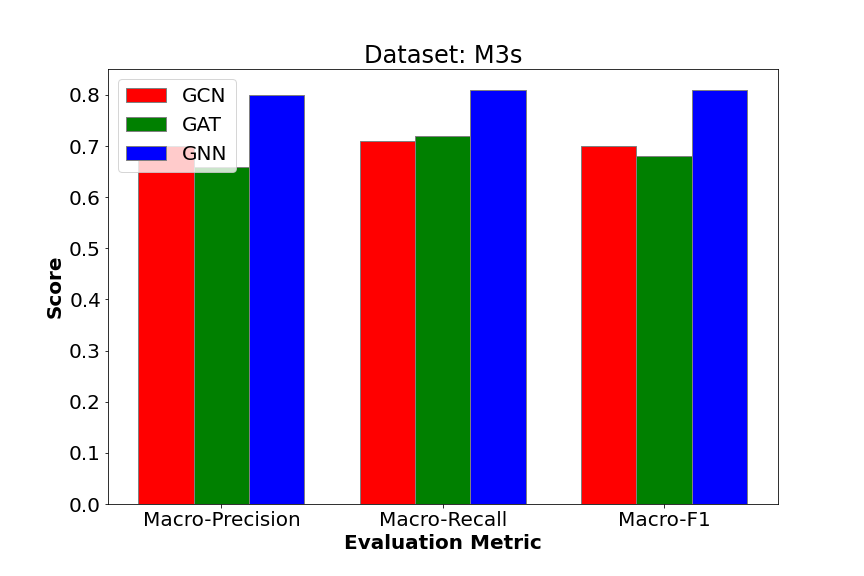}
    \caption{M3s Dataset}
    \label{fig:fig1}
  \end{subfigure}%
  \begin{subfigure}[b]{0.5\linewidth}
    \centering\includegraphics[scale = 0.2]{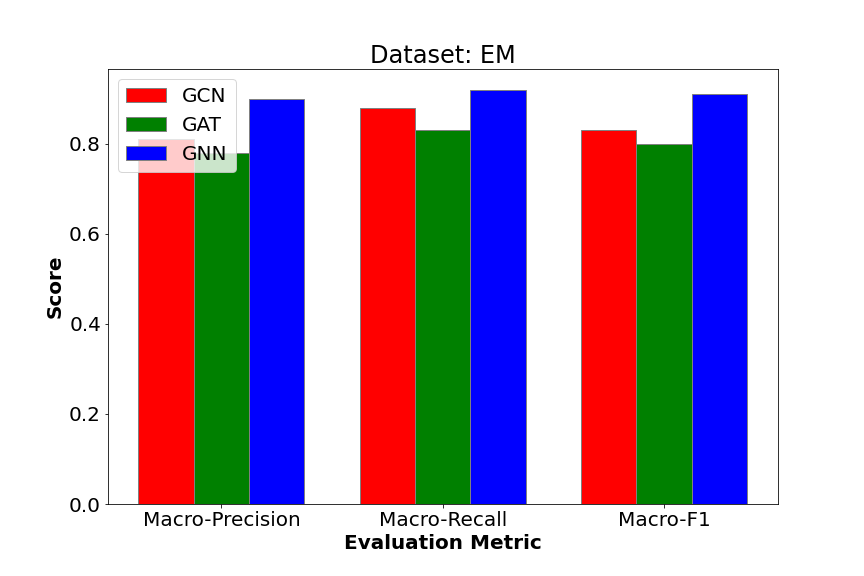}
    \caption{EM Dataset}
    \label{fig:fig2}
  \end{subfigure}
  \caption{Results of the various Graph Neural Network architectures on the test set of the datasets}
  \label{fig:result-1}
\end{figure}

\begin{itemize}
    \item Initially, to validate our proposed Relevant Node Detection model, we compare it against other variants of Graph Neural Network. We use the same architecture of the model, but only change the GNN layers with Graph Convolutional Neural Network \cite{kipf2017semi:gcn}, and Graph Attention Network \cite{velickovic2018graph:gat} for comparison. We use macro-average precision, recall, and F1-score as the metrics for evaluation.
    
    \item To compare the performance of the GrabQC generated contextual query, we use an Information Retrieval (IR) based evaluation. We keep the IR system as constant and compare the performance by providing the various baseline queries. We use Recall@$k$ as the evaluation metric, as we are concerned with the method's ability to retrieve the specific code from the large pool of labels. The retrieved codes can be pruned further using rules or business logic.
    
    \item As GrabQC involves machine-learning for generating the contextual query, we compare the performance against other ML-based methods used for autocoding. We compare against other Neural network architectures (CNN, Bi-GRU) and two popular methods (CAML, MultiResCNN) in the literature \cite{mullenbach-etal-2018-explainable:caml,li2020multirescnn}. We use label macro-average precision, recall, F1-score, and also Recall@$k$ as the evaluation metrics. We use macro-average as it reflects the performance on a large label set and emphasises rare-label prediction \cite{mullenbach-etal-2018-explainable:caml}.
    
\end{itemize}

\subsection{Implementation Details} \label{implementation}
We use UMLS\footnote{https://www.nlm.nih.gov/research/umls/index.html} and SNOMED\footnote{https://www.snomed.org/} as Knowledge Base ($\mathcal{B}$) for the entity linking (Section \ref{Named Entity Recognition}) and path extraction. We use SciSpacy \cite{neumann-etal-2019-scispacy} for Named Entity Recognition and Entity Linking to UMLS. We use ElasticSearch\footnote{https://www.elastic.co/} in its default settings as our IR system. We index the IR system by the ICD codes and their corresponding description. There are certain keywords (\textbf{External Keywords}) that occur in the ICD code descriptions but never occur in the clinical notes (e.g., without, unspecified). We add the Top 4 occurring external keywords as nodes to the $\mathcal{G}$ and link them with the linked entities in the query. We use PyTorch for implementing the relevant node detection model.

\section{Results and Discussion} \label{Results}


\subsection{Performance of the Relevant Node Detection Model} \label{result:1}

In the Contextual Graph, the query's entities are the centre nodes that link the other entities. Hence the information of the centre node is essential for classifying whether the node is relevant or not. Since the GNN model captures that information explicitly by skip connections and without normalizing on the neighbors, it performs better than the other compared methods seen in Fig. \ref{fig:result-1}.   

\subsection{Query-level Comparison} \label{result:2}

\renewcommand{\arraystretch}{1.2}
\begin{table}[t]
\centering

\resizebox{\textwidth}{!}{%
\begin{tabular}{@{}llccc@{}}
\toprule
\textbf{Dataset} & \textbf{Query Type} & \multicolumn{1}{l}{\textbf{Recall @ 1}} & \multicolumn{1}{l}{\textbf{Recall @ 8}} & \multicolumn{1}{l}{\textbf{Recall @ 15}} \\ \midrule
\textbf{M3s} & Normal & 0.2098 & 0.5830 & 0.6684 \\
\multicolumn{1}{c}{} & Contextual Graph & 0.1911 & 0.5540 & 0.6404 \\
\multicolumn{1}{c}{} & GrabQC & \textbf{0.2130} & \textbf{0.5978} & \textbf{0.6767} \\
\multicolumn{1}{c}{} & Normal + External & 0.3000 & 0.5732 & 0.6242 \\
\multicolumn{1}{c}{} & Contextual Graph + External & 0.2792 & 0.5839 & 0.6550 \\
\multicolumn{1}{c}{} & GrabQC + External & \textbf{0.3129*} & \textbf{0.6230*} & \textbf{0.6876*} \\

\midrule

\textbf{EM} & Normal & \textbf{0.2562} & \textbf{0.5999} & 0.6541 \\
 & Contextual Graph (CG) & 0.2474 & 0.5888 & 0.6462 \\
 & GrabQC & 0.2525 & 0.5996 & \textbf{0.6547} \\
 & Normal + External & 0.2752 & 0.5539 & 0.5999 \\
 & CG + External & 0.2472 & 0.5941 & 0.6504 \\
 & GrabQC + External & \textbf{0.3437*} & \textbf{0.6465*} & \textbf{0.6872*} \\ \bottomrule
\end{tabular}%
}
\caption{Results of the different queries along with the contextual query generated by GrabQC. Here, Normal refers to the query ($q$) extracted from the clinical text; Contextual Graph refers to the $\mathcal{G}$,  External refers to the concatenation of the described query with the external keywords described in Section \ref{implementation}.  The best score, with the external keywords, is marked in \textbf{bold*} and without the external keywords is marked in \textbf{bold}.}
\label{tab:query-comparison}
\end{table}

From Table \ref{tab:query-comparison}, it is evident that the performance of the system changes on the addition of keywords from the source document. The system's performance degrades when we pass $\mathcal{G}$ as the query. This behavior validates the fact that there are many non-relevant terms in the $\mathcal{G}$ that makes the IR system provide noisy labels. When GrabQC filters the relevant terms, the performance increases substantially and is better than the extracted query. Also, the addition of external keywords significantly improves the performance, as the IR system is indexed using the ICD-code descriptions. Our method GrabQC again filters relevant terms from the external keywords, which is evident from the increase in performance compared to the standard query with external keywords.

\subsection{Comparison with Machine Learning Models}
\label{result:3}

\begin{table}[]
\centering
\resizebox{\textwidth}{!}{%
\begin{tabular}{@{}ll|lll|ll@{}}
\toprule
\multicolumn{1}{l}{\textbf{Dataset}} & \multicolumn{1}{l|}{\textbf{Method}} & \multicolumn{3}{c|}{\textbf{Macro}} & \multicolumn{2}{c}{\textbf{Recall}} \\
\multicolumn{1}{l}{} & \multicolumn{1}{l|}{} & Precision & Recall & \multicolumn{1}{l|}{F1-score} & @8 & @15 \\ \midrule
\textbf{M3s} & CNN & 0.0310 & 0.0273 & 0.0291 & 0.5177 & 0.5783 \\
 & Bi-GRU & 0.0222 & 0.0104 & 0.0141 & 0.4578 & 0.5079 \\
 & CAML \cite{mullenbach-etal-2018-explainable:caml} & 0.0093 & 0.0067 & 0.0078 & 0.2383 & 0.3064 \\
 & MultiResCNN \cite{li2020multirescnn} & 0.0118 & 0.0071 & 0.0089 & 0.4701 & 0.5415 \\
 & GrabQC & 0.1164 & 0.1153 & 0.1027 & 0.5978 & 0.6767\\
 & GrabQC + External & \textbf{0.1334} & \textbf{0.1193} & \textbf{0.1129} & \textbf{0.6230} & \textbf{0.6876} \\
 
\midrule

\textbf{EM} & CNN & 0.0711 & 0.0538 & 0.0613 & 0.6459 & 0.6771 \\
 & Bi-GRU & 0.0755 & 0.0431 & 0.0549 & 0.6333 & 0.6731 \\
 & CAML \cite{mullenbach-etal-2018-explainable:caml} & 0.0267 & 0.0158 & 0.0199 & 0.1410 & 0.2040 \\
 & MultiResCNN \cite{li2020multirescnn} & 0.0288 & 0.0176 & 0.0218 & 0.1410  & 0.2057  \\
 & GrabQC & 0.1333 & 0.1266 & 0.1226 & 0.5996 & 0.6547\\
 & GrabQC + External & \textbf{0.1670} & \textbf{0.1567} & \textbf{0.1534} & \textbf{0.6465} & \textbf{0.6872} \\ \bottomrule
\end{tabular}%
}
\caption{Experimental results of the proposed method along with the other machine learning baselines on the test set. The best scores are marked in \textbf{bold}.}
\label{tab:ml-results}
\end{table}

From Table \ref{tab:ml-results}, we observe that the macro-average performance of the methods is deficient. The deficiency is due to the high number of labels in the ICD coding task. Our method of using the contextual query generated by GrabQC performs better than the other compared methods. Complex models like CAML and MultiResCNN are unable to perform well in this case due to a large number of labels and a lesser number of training samples. Even with a limited amount of training data, our method can perform significantly better than the other baselines.

\subsection{Analysis}

We analyze the effect of the number of layers in the Graph Neural Network on the performance of GrabQC. The results of this analysis are tabulated in Table \ref{tab:layers}. We observe that choosing the number of layers as 3 gave us the best result in both the datasets.

\begin{table}[H]
\parbox{.35\linewidth}{
\centering
\resizebox{0.25\textwidth}{!}{%
\begin{tabular}{@{}ccc@{}}
\toprule
\begin{tabular}[c]{@{}l@{}} \# layers \\ in GNN\end{tabular} & M3s & EM \\ \midrule
1 & 0.89 & 0.78 \\
2 & 0.90 & 0.80 \\
3 & \textbf{0.91} & \textbf{0.80} \\
4 & 0.90 & 0.79 \\ \bottomrule
\end{tabular}%
}
\caption{Macro-average F1-score on the test set of the datasets obtained by varying the number of layers in the Graph Neural Network.}
\label{tab:layers}
}
\hfill
\parbox{.6\linewidth}{
\centering
\resizebox{0.6\textwidth}{!}{%
\begin{tabular}{ll}
\hline
\textbf{Normal Query ($q$)} & \textbf{Contextual Query ($q_c$)} \\ \hline
Sciatica & \begin{tabular}[c]{@{}l@{}}Sciatica pain right \\ lower back\end{tabular} \\
Acute knee pain & Acute knee pain left \\
Laceration & Lip Laceration \\
Strain of lumbar region & \begin{tabular}[c]{@{}l@{}}Strain of lumbar region\\ lower back pain\end{tabular} \\
History of breast cancer & \begin{tabular}[c]{@{}l@{}}History of breast cancer \\ left breast carcinoma\end{tabular} \\ \hline
\end{tabular}%
}
\caption{Examples of GrabQC contextual queries generated for the queries extracted from the clinical note.}
\label{tab:qc-examples}
}
\end{table}

Table \ref{tab:qc-examples} tabulates the contextual query generated by GrabQC. From the examples, we see that the GrabQC method can contextualize the standard query. GrabQC can associate the extracted query with the region of the diagnosis, the laterality, and associated conditions. The contextual query can also serve as an explanation for the ICD codes returned by the IR system. 

\section{Conclusion} \label{Conclusion}

We proposed an automation process that expedites the current medical coding practice in the industry by decreasing the turn-around time. Our method mimics the manual coding process that provides full explainability. We proposed \textbf{GrabQC}, an automatic way of generating a query from clinical notes using a graph neural network-based query contextualization module. We successfully demonstrated the effectiveness of the proposed method in three settings. As a next step, we plan to use Reinforcement Learning (RL) for automatically contextualizing the query, without the need for relevance labels. The proposed contextualization method can also be extended as hard attention for DL-based methods. 

\subsubsection{Acknowledgements}

The authors would like to thank Buddi.AI for funding this research work through their project RB1920CS200BUDD008156.

%
%
%
\bibliographystyle{splncs04}
\bibliography{ref.bib}

\begin{thebibliography}{10}
\providecommand{\url}[1]{\texttt{#1}}
\providecommand{\urlprefix}{URL }
\providecommand{\doi}[1]{https://doi.org/#1}

\bibitem{Bhatia16}
Bhatia, K., Dahiya, K., Jain, H., Mittal, A., Prabhu, Y., Varma, M.: The
  extreme classification repository: Multi-label datasets and code (2016),
  \url{http://manikvarma.org/downloads/XC/XMLRepository.html}

\bibitem{DUARTE201864:DLforICD}
Duarte, F., Martins, B., Pinto, C.S., Silva, M.J.: Deep neural models for
  icd-10 coding of death certificates and autopsy reports in free-text. Journal
  of Biomedical Informatics  \textbf{80},  64 -- 77 (2018).
  \doi{https://doi.org/10.1016/j.jbi.2018.02.011},
  \url{http://www.sciencedirect.com/science/article/pii/S1532046418300303}

\bibitem{article:25-billion}
Farkas, R., Szarvas, G.: Automatic construction of rule-based icd-9-cm coding
  systems. BMC bioinformatics  \textbf{9 Suppl 3}, ~S10 (02 2008).
  \doi{10.1186/1471-2105-9-S3-S10}

\bibitem{inproceedings:ml-3}
Karimi, S., Dai, X., Hassanzadeh, H., Nguyen, A.: Automatic diagnosis coding of
  radiology reports: A comparison of deep learning and conventional
  classification methods (08 2017). \doi{10.18653/v1/W17-2342}

\bibitem{inproceedings:text-summarization-for-ICD}
Kavuluru, R., Han, S., Harris, D.: Unsupervised extraction of diagnosis codes
  from emrs using knowledge-based and extractive text summarization techniques.
  vol.~7884 (05 2013). \doi{$10.1007/978-3-642-38457-8_7$}

\bibitem{article:adam}
Kingma, D., Ba, J.: Adam: A method for stochastic optimization. International
  Conference on Learning Representations  (12 2014)

\bibitem{kipf2017semi:gcn}
Kipf, T.N., Welling, M.: Semi-supervised classification with graph
  convolutional networks. In: International Conference on Learning
  Representations (ICLR) (2017)

\bibitem{article:ml-2}
Koopman, B., Zuccon, G., Nguyen, A., Bergheim, A., Grayson, N.: Automatic
  icd-10 classification of cancers from free-text deathcertificates.
  International journal of medical informatics  \textbf{84} (08 2015).
  \doi{10.1016/j.ijmedinf.2015.08.004}

\bibitem{10.1145/243199.243276:ml-feature}
Larkey, L.S., Croft, W.B.: Combining classifiers in text categorization. In:
  Proceedings of the 19th Annual International ACM SIGIR Conference on Research
  and Development in Information Retrieval. p. 289–297. SIGIR '96,
  Association for Computing Machinery, New York, NY, USA (1996).
  \doi{10.1145/243199.243276}, \url{https://doi.org/10.1145/243199.243276}

\bibitem{li2020multirescnn}
Li, F., Yu, H.: Icd coding from clinical text using multi-filter residual
  convolutional neural network. In: Proceedings of the Thirty-Fourth AAAI
  Conference on Artificial Intelligence (2020)

\bibitem{inproceedings:additional-info}
Lima, L., Laender, A., Ribeiro-neto, B.: A hierarchical approach to the
  automatic categorization of medical documents. pp. 132--139 (01 1998).
  \doi{10.1145/288627.288649}

\bibitem{10.5555/1690219.1690287:distant-supervision}
Mintz, M., Bills, S., Snow, R., Jurafsky, D.: Distant supervision for relation
  extraction without labeled data. In: Proceedings of the Joint Conference of
  the 47th Annual Meeting of the ACL and the 4th International Joint Conference
  on Natural Language Processing of the AFNLP: Volume 2 - Volume 2. p.
  1003–1011. ACL '09, Association for Computational Linguistics, USA (2009)

\bibitem{Morris_Ritzert_Fey_Hamilton_Lenssen_Rattan_Grohe_2019:GNN}
Morris, C., Ritzert, M., Fey, M., Hamilton, W.L., Lenssen, J.E., Rattan, G.,
  Grohe, M.: Weisfeiler and leman go neural: Higher-order graph neural
  networks. Proceedings of the AAAI Conference on Artificial Intelligence
  \textbf{33}(01),  4602--4609 (Jul 2019). \doi{10.1609/aaai.v33i01.33014602},
  \url{https://ojs.aaai.org/index.php/AAAI/article/view/4384}

\bibitem{mullenbach-etal-2018-explainable:caml}
Mullenbach, J., Wiegreffe, S., Duke, J., Sun, J., Eisenstein, J.: Explainable
  prediction of medical codes from clinical text. In: Proceedings of the 2018
  Conference of the North {A}merican Chapter of the Association for
  Computational Linguistics: Human Language Technologies, Volume 1 (Long
  Papers). pp. 1101--1111. Association for Computational Linguistics, New
  Orleans, Louisiana (Jun 2018). \doi{10.18653/v1/N18-1100},
  \url{https://www.aclweb.org/anthology/N18-1100}

\bibitem{neumann-etal-2019-scispacy}
Neumann, M., King, D., Beltagy, I., Ammar, W.: {S}cispa{C}y: {F}ast and
  {R}obust {M}odels for {B}iomedical {N}atural {L}anguage {P}rocessing. In:
  Proceedings of the 18th BioNLP Workshop and Shared Task. pp. 319--327.
  Association for Computational Linguistics, Florence, Italy (Aug 2019).
  \doi{10.18653/v1/W19-5034}, \url{https://www.aclweb.org/anthology/W19-5034}

\bibitem{article-IR2}
Park, H., Castaño, J., Ávila Williams, M., Perez, D., Berinsky, H., Gambarte,
  M., Luna, D., Otero, C.: An information retrieval approach to icd-10
  classification. Studies in health technology and informatics  \textbf{264},
  1564--1565 (08 2019). \doi{10.3233/SHTI190536}

\bibitem{inproceedings:IR}
Rizzo, S., Montesi, D., Fabbri, A., Marchesini, G.: Icd code retrieval: Novel
  approach for assisted disease classification. pp. 147--161 (07 2015).
  \doi{$10.1007/978-3-319-21843-4_12$}

\bibitem{article:DLforICD2}
Shi, H., Xie, P., Hu, Z., Zhang, M., Xing, E.: Towards automated icd coding
  using deep learning  (11 2017)

\bibitem{10.3389/fbioe.2020.00867:KGforICD}
Teng, F., Yang, W., Chen, L., Huang, L., Xu, Q.: Explainable prediction of
  medical codes with knowledge graphs. Frontiers in Bioengineering and
  Biotechnology  \textbf{8}, ~867 (2020). \doi{10.3389/fbioe.2020.00867},
  \url{https://www.frontiersin.org/article/10.3389/fbioe.2020.00867}

\bibitem{velickovic2018graph:gat}
Veli{\v{c}}kovi{\'{c}}, P., Cucurull, G., Casanova, A., Romero, A., Li{\`{o}},
  P., Bengio, Y.: {Graph Attention Networks}. International Conference on
  Learning Representations  (2018),
  \url{https://openreview.net/forum?id=rJXMpikCZ}, accepted as poster

\bibitem{inproceedings:ml-5}
Vu, T., Nguyen, D.Q., Nguyen, A.: A label attention model for icd coding from
  clinical text. pp. 3307--3313 (07 2020). \doi{10.24963/ijcai.2020/457}

\bibitem{inproceedings:25-billion-2}
Xie, P., Xing, E.: A neural architecture for automated icd coding. pp.
  1066--1076 (01 2018). \doi{10.18653/v1/P18-1098}

\bibitem{unknown:bertXML}
Zhang, Z., Liu, J., Razavian, N.: Bert-xml: Large scale automated icd coding
  using bert pretraining  (05 2020)

\end{thebibliography}

\end{document}